\documentclass[
twocolumn,
]{ceurart}

\usepackage{listings}
\begin{document}

\copyrightyear{2022}
\copyrightclause{Copyright for this paper by its authors.
  Use permitted under Creative Commons License Attribution 4.0
  International (CC BY 4.0).}

\conference{STRL'22: First International Workshop on Spatio-Temporal Reasoning and Learning,
  July 24, 2022, Vienna, Austria}

\title{Scene Separation \& Data Selection: Temporal Segmentation Algorithm for Real-Time Video Stream Analysis}


\author[1,2]{Yuelin Xin}[%
orcid=0000-0002-9732-2414,
email=sc20yx2@leeds.ac.uk,
]
\cormark[1]
\fnmark[1]
\address[1]{Southwest Jiaotong University, Chengdu, China}
\address[2]{University of Leeds, Leeds, UK}

\author[1,2]{Zihan Zhou}[%
orcid=0000-0003-2613-7569,
email=sc20zz2@leeds.ac.uk,
]
\fnmark[1]

\author[1,2]{Yuxuan Xia}[%
orcid=0000-0002-1185-2722,
email=sc202yx@leeds.ac.uk,
]
\fnmark[1]

\cortext[1]{Corresponding author.}
\fntext[1]{These authors contributed equally.}

\begin{abstract}
  We present 2SDS (Scene Separation and Data Selection algorithm), a temporal segmentation algorithm used in real-time video stream interpretation. It complements CNN-based models to make use of temporal information in videos. 2SDS can detect the change between scenes in a video stream by com-paring the image difference between two frames. It separates a video into segments (scenes), and by combining itself with a CNN model, 2SDS can select the optimal result for each scene.
In this paper, we will be discussing some basic methods and concepts behind 2SDS, as well as presenting some preliminary experiment results regarding 2SDS. During these experiments, 2SDS has achieved an overall accuracy of over 90
\end{abstract}


\maketitle

\section{Introduction}
Image recognition models have gone increasingly accurate in the past few years, yet video semantics tasks are still challenging. A detailed comprehension on video stream could play a significant part in video accessibility \cite{DBLP:journals/expert/StappenBCSC21}, surveillance footage auto-interpretation \cite{DBLP:journals/mta/PatelV0O22,DBLP:journals/csur/PalSDKRP21}, and so on. These technologies have already been proven useful on large video platforms like YouTube, used for real-time video interpretation and video topic analysis.
\subsection{The Problem}
In the processing of video stream, a 2D CNN can be extended into 3D CNN by adding a temporal dimension \cite{DBLP:journals/corr/abs-1711-08200}, but this approach can be hazardous if the video is too long, or it is of indefinite length. However, a 2D CNN is still very usable in a traditional image recognition or image segmentation task.

\begin{figure}
  \centering
  \includegraphics[width=\linewidth]{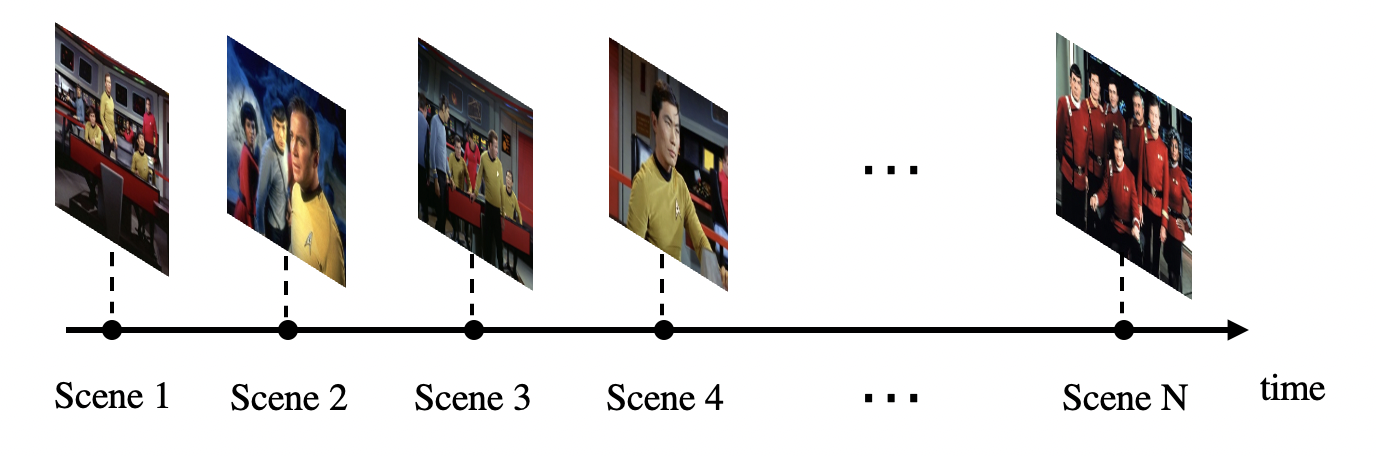}
  \caption{\textbf{Overall effect of the scene separation procedure.} The whole video stream will be separated into scenes, in each of which the images in the video remain relatively stationary.}\label{fig:scene}
\end{figure}

The problem is that 2D CNNs only recognise a video as discrete images, rather than a continuous stream of images. This poses some issues. For example, a CNN model could not resolve the motion of a person (e.g., walking, dancing) be-cause the person is stationary in every frame, and this will cause the 
loss of significant information in video analysis. So, we need to devise an implementation that complements the CNN model to solve the continuity issue. This implementation should group the discrete frames (adjacent on the temporal axis) that look similar to each other into \emph{scenes}, this procedure is what we call \emph{temporal segmentation} (also referred as \emph{scene separation} in 2SDS, see Fig.~\ref{fig:scene} for example).

\subsection{Related Work}
\textbf{SlowFast Networks.} The SlowFast Networks use a two-pathway architecture for video recognition, the slow pathway (low frame rate) is used to capture spatial semantics, and the fast pathway (high frame rate) is used to capture temporal semantics like motions in a relatively fine temporal resolution \cite{DBLP:conf/iccv/Feichtenhofer0M19}.

\subsection{Our Work}
\begin{figure}
  \centering
  \includegraphics[width=\linewidth]{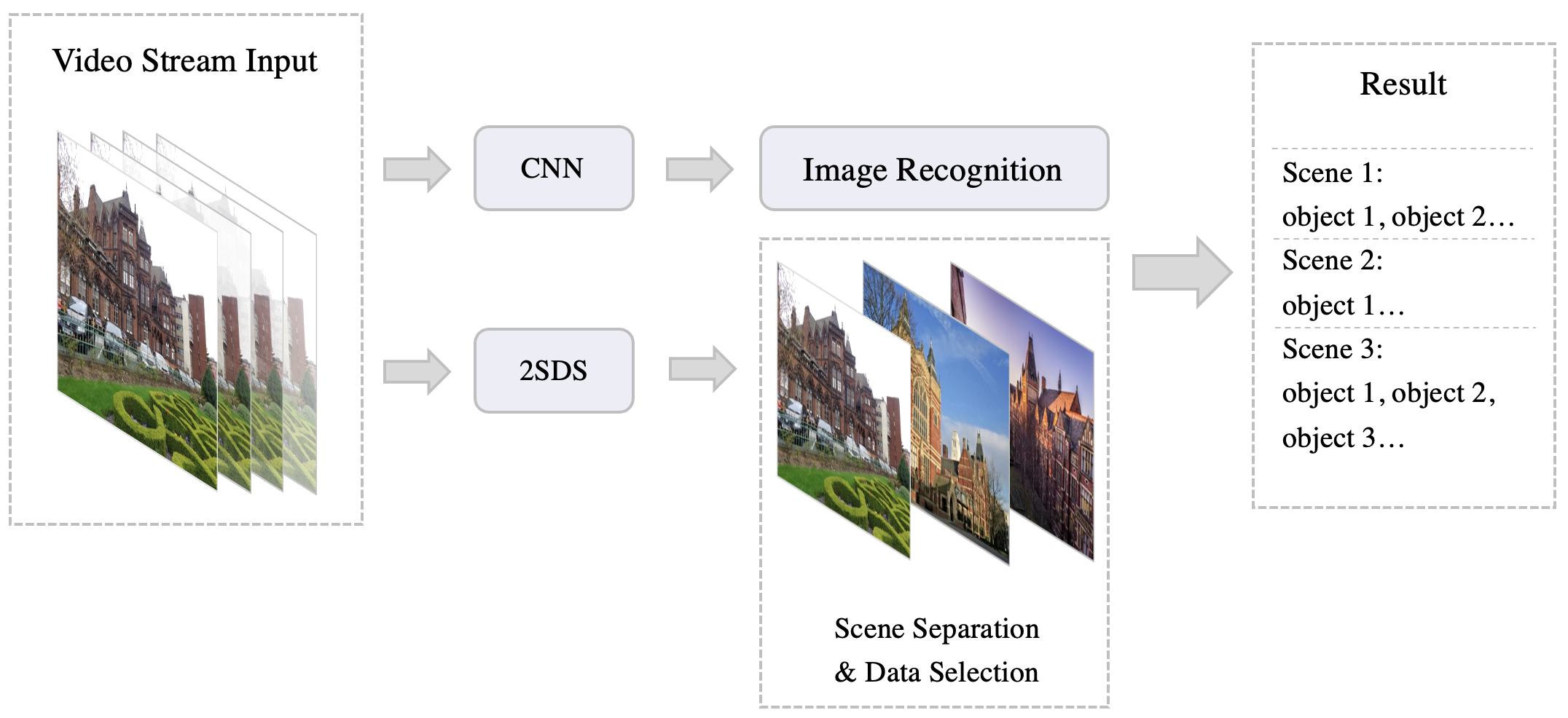}
  \caption{\textbf{2SDS used together with a CNN model.} The 2SDS algorithm can separate the scenes in a continuous video stream and select the result produced by the CNN model, the two together, can produce a scene-separated recognition result.}\label{fig:model}
\end{figure}

What we have achieved is to devise the temporal segmentation algorithm, 2SDS, which stands for “Scene Separation and Data Selection algorithm”. It can slice the video stream into segments on the temporal axis, so it can be interpreted using 2D CNN models while preserving critical information on the temporal dimension. By combining 2SDS with a CNN model (Fig.~\ref{fig:model}), this implementation is similar to the SlowFast Networks on splitting the input into two pathways, in which the 2SDS is similar to the fast pathway of the SlowFast Networks, except we do not introduce another neural network, but we replace the network with the faster 2SDS, which guarantees even better temporal resolution.

\section{Motivation: Why Not Neural Networks}
Traditionally, RNN-based models have been quite successful in processing sequential information like time. However, the usage of RNN or even neural networks is not practical in time sensitive tasks like real-time object recognition and live video stream analysis, which requires fast responding algorithms, and RNNs usually cannot meet those requirements.

RNN-based models like LSTM \cite{DBLP:journals/neco/HochreiterS97} generally have a longer respond time even compared to CNN-based models (although the difference between them vary with different settings of hyperparameters). A CNN + RNN architecture model would mean the doubling of processing time, which is something we would rather avoid when dealing with video stream analysis tasks.

However, RNNs do have the advantage of acting upon temporal information, especially for models like LSTM. So, we need to help the CNN-based models to preserve temporal information, and that is where we introduce our temporal segmentation algorithm, 2SDS.

By adding the 2SDS algorithm, alongside a CNN model, we were able to achieve RNN-like results. In the meantime, by avoiding the introduction of a neural network, this implementation is also faster than the CNN + RNN architecture or the CNN + CNN architecture.

\section{Method: 2SDS}
2SDS stands for “Scene Separation and Data Selection algorithm”. It works as a temporal segmentation and result selection algorithm to complement CNN-based models. It contains a two-part procedure of separating the video stream into segments and selecting a representative recognition result from the CNN model for output.

2SDS utilises the difference hash (dHash) method \cite{Krawetz} to obtain the rough image difference between two frames, if two frames have a very little difference, they will be grouped into the same scene. This method involves a few simple steps to calculate, and it is the most important method 2SDS uses to achieve scene separation. As the calculation is 
relatively simple and straight forward, this makes 2SDS extremely fast on scene separation.

Also, 2SDS uses a pooling-layer-like data smoothing and data selection method to pick out the representative recognition result for a particular scene. This method can generally improve the accuracy of the output because it can smooth out the data on undesired frame moving (e.g., camera shaking, broken frames). Alongside the data smoothing mechanism, another data selection mechanism is implemented to select the representative recognition result (referred as \emph{representative} in the following sections) from the whole data segment of a scene (referred as \emph{candidate} in the following sections).

\subsection{Scene Separation: based on dHash}
\begin{figure}
  \centering
  \includegraphics[width=\linewidth]{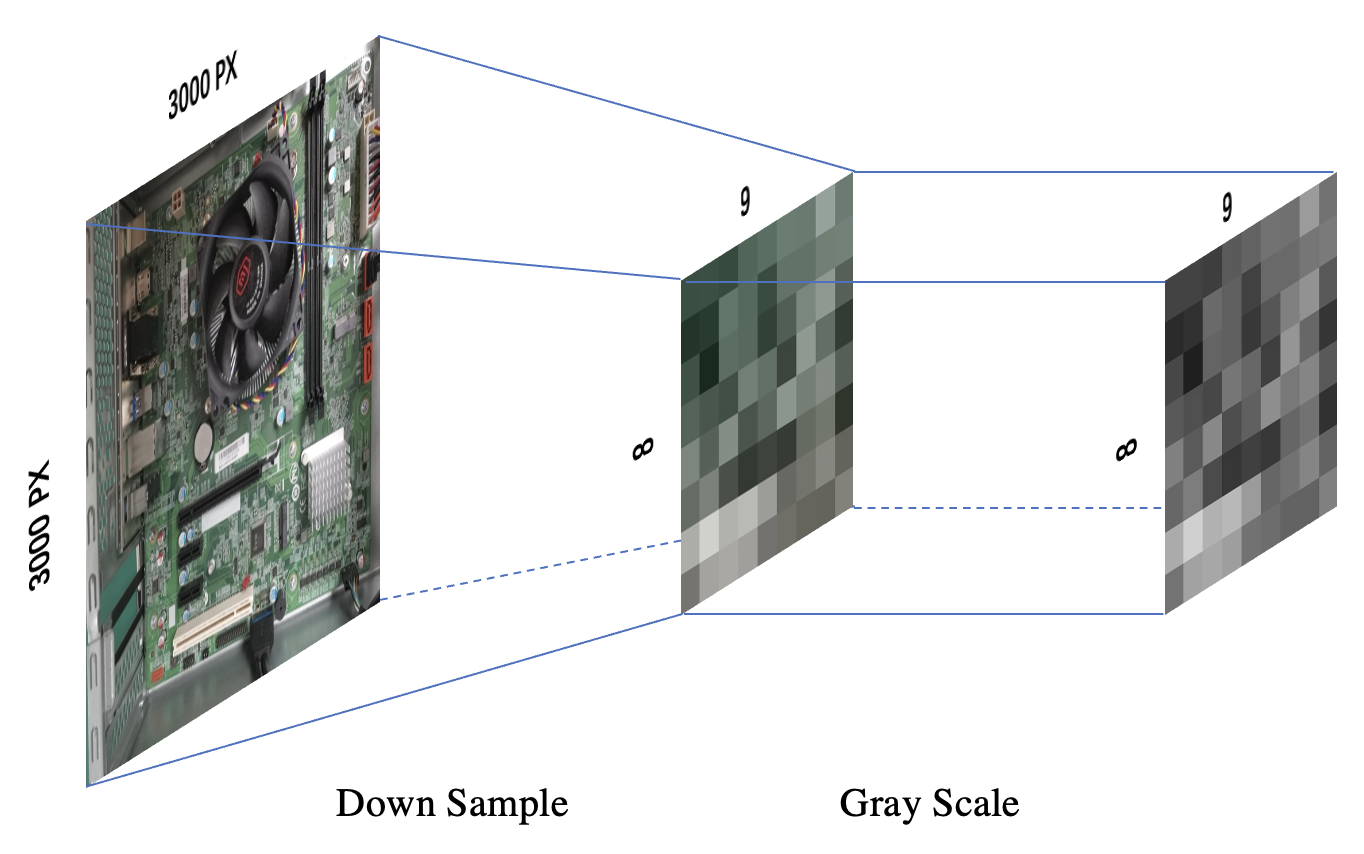}
  \caption{\textbf{Image processing in 2SDS based on an improved dHash algorithm.} The two image processing parts in 2SDS, the first step is down sample, and the second step is gray scale conversion.}\label{fig:procedure}
\end{figure}

The scene separation procedure of 2SDS (Fig.~\ref{fig:procedure}) is based on an improved version of the dHash algorithm, which is originally used to judge the similarity of two images. By applying the scene separation procedure, the temporal information can be preserved by the sequencing of the separated scenes. The exact workflow of the scene separation process is discussed extensively below.

\textbf{Down sampling.} To make a rough comparison between two frames in a video, the frames need to be down sampled from their original size to an 8 by 9 (row by column) sub-image. This approach can both simplify the remaining calculation and make the algorithm less sensitive to subtle changes between frames.

\textbf{Gray scale.} We apply gray scale manipulation on the previous sub-image using the Luminosity algorithm, this step is purely for reducing the complexity of calculating the difference on 3 channels. By converting the RGB channels into one gray scale channel, this approach dramatically lessens the complexity of the algorithm.

\begin{figure}
  \centering
  \includegraphics[width=0.5\linewidth]{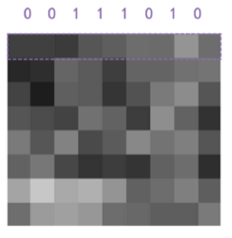}
  \caption{Example on binary sequence conversion. The derived binary sequence of row 1 in this case is 00111010.}\label{fig:sequence}
\end{figure}

\textbf{Calculate Hash value.} The derived gray scale image is converted into a single 16-bit hexadecimal hash value. The algorithm looks at all the 8 rows separately, each row has 9 gray scale values from 0 to 255. These 9 values are converted to 8 binary numbers under the following rules:
\begin{enumerate}
  \item[(a)] One binary value stands for the gray scale difference between two adjacent pixels.
  \item[(b)] If the gray scale value of the pixel on the left is greater than the pixel on the right, the binary value should be 1, otherwise, it should be 0.
  \item[(c)] Every row should end up with an 8-bit long binary sequence.
\end{enumerate}
An example is given in Fig.~\ref{fig:sequence}.

Using this method, we can derive eight 8-bit long binary sequences, each of them can be represented by a 2-bit long hexadecimal value. And by concatenating all the 2-bit hexadecimal values, we can obtain a 16-bit long hexadecimal hash value, and this value will represent the whole image (this is also the reason why the original image is down sampled into an 8 by 9 sub-image rather than an 8 by 8 sub-image, because the 8 by 8 image will face some inconvenience when converting into a hexadecimal hash value). 

\textbf{Calculating the Hamming distance.} By calculating the Hamming distance between the hash values of two adjacent frames, we can judge whether the two frames are in the same scene or not. If the Hamming distance is greater than a threshold (usually 5), we consider the two frames to be in two different scenes, and we can separate them accordingly. For the calculation of Hamming distance, we can simply use an Exclusive Or operator on the two hash values, here is an example below (the Hamming distance is 7 in this case):
\begin{equation}
  \text{c4e0d8988c989898}\ \oplus\ \text{eee6989c8c989898}\ =\ \ 7
\end{equation}

\subsection{Data Selection and Data Smoothing}
When the scene separation process detected a new scene, the data collected on the previous scene is packed into an array. This array contains all the recognition results produced by the CNN model in the previous scene, and the CNN model would have a recognition output on every frame in this scene.

To have a solid output for 2SDS, we need to perform 2 extra steps: data smoothing and data selection. The method that is implemented here is inspired by the pooling layer in a convolutional neural network.

\textbf{Data smoothing procedure.} This step is also called LWAP (Length Weighted Average Pooling). We start by segmenting the array containing all the recognition data into small groups of a defined size. Then, we apply the following formulas:
\begin{equation}
  WAL=\frac{\sum_{i=1}^{i\le\varphi}{{(L}_i\times\omega_i)}}{\sum_{i=1}^{i\le\varphi}{\ \omega}_i}
\end{equation}
\begin{equation}
  \left\{\begin{array}{l}
    f_i(D)=\min_{i=0}{|\operatorname{card}(D_i)-\operatorname{WAL}|}\\
    C_{I}=[c\in D\ |\ \operatorname{card}(c)=f_i(D)]
  \end{array}\right.
\end{equation}
Here, $L_i$ stands for the length of each recognition data, for example, in “object 1, object 2, object 1”, $L_i=3$. $\omega_i$ stands for the weight of each recognition data which is $0.1\times L_i$. $\operatorname{card}(D_i)$ stands for the cardinality of set  $D_i$, where $D_i$ is the segments previously obtained by segmenting the original array.

This approach is inspired by the pooling layer in CNN, but instead of a Max Pooling operation, the data smoothing procedure uses a Weighted Average Pooling operation. By using this data smoothing procedure, we can avoid unwanted recognition results like broken frames or camera flashes.

\textbf{Data selection procedure.} We apply a data selection procedure that uses the similar approach that we previously used in the data smoothing procedure, which is also a Weighted Average Pooling operation. This procedure will select the recognition result from a frame whose feature intensity is the closest to the weighted average value of all the candidates (feature intensity refers to the number of different classes of objects in a particular frame).

Finally, we can output the result that we obtained in the previous steps as the representative of the whole scene. This particular recognition result will be used to represent the whole scene it is located in, and through the help of NLP and other models, this can even be used to output the natural language interpretation of this video scene.

\section{Experiments}
Due to the lack of similar algorithms and datasets, we could only provide some preliminary and experimental usage of the 2SDS algorithm\footnote{We only did some preliminary experiments on the accuracy of 2SDS on scene separation (temporal segmentation) tasks, more detailed experiments are still needed to be conducted.}.

We choose YOLOv5s as our image recognition CNN for this experiment, and we have built an experimental dataset on video object detection using selected YouTube videos in the YouTube-VOS dataset \cite{DBLP:journals/corr/abs-1809-03327}. Although the YOLOv5s algorithm is trained on the COCO dataset, this CNN model is still sufficiently usable in this experiment for it is not the key focus of this experiment.

We are most interested in how 2SDS will perform in scene separation (temporal segmentation) tasks. We classified the testing videos into 3 classes: interviews, vibrant, and hybrid. 

The interviews are usually straight forward and easier to undergo scene separation tasks. Vibrant videos are the more difficult ones due to their fast-moving images and transition effects that might seem deceptive to 2SDS. The hybrid video sits in between the first two types, they have some features of the interview videos, as well as features from the vibrant videos, their difficulty should sit in the middle.

\subsection{Interview Video Tests}
We conducted 3 separate experiments using interview videos (Table~\ref{tab:interview}). The total amount of scenes in these 3 experiments is $82$. The overall accuracy of 2SDS during these experiments is $90.10\%$. There are 2 cases where we find the 2SDS algorithm actually over-judged the transition between two scenes. This is potentially a sensitivity issue posed by the hard coded threshold during scene separation.

\begin{table}
  \caption{Interview video tests results.}
  \label{tab:interview}
  \begin{tabular}{lcr}
    \toprule
    Experiment No.&	Output - Truth&	Accuracy\\
    \midrule
    Interview 1 &	25 - 25 &	$100.00\%$\\
Interview 2 &	35 - 29 &	$82.86\%$\\
Interview 3 &	31 - 28 &	$90.32\%$\\
  \bottomrule
\end{tabular}
\end{table}

\subsection{Vibrant Video Tests}
We conducted 2 separate experiments using vibrant videos (Table~\ref{tab:vibrant}). The total amount of scenes in these 2 experiments is $51$. The overall accuracy of 2SDS during the two experiments is $54.90\%$. The accuracy in vibrant videos is substantially lower than interview videos for the 2SDS is unable to separate two fast-moving scenes effectively. It is important to notice that we used harsh videos like sport videos and dynamic advertisement videos in this experiment, so the performance of the 2SDS is expected to be much lower comparing to the previous experiment. This is the biggest limitation of 2SDS, but this issue is addressable with future improvements of the algorithm.

\begin{table}
  \caption{Vibrant video tests results.}
  \label{tab:vibrant}
  \begin{tabular}{lcr}
    \toprule
    Experiment No.&	Output - Truth&	Accuracy\\
    \midrule
    Vibrant 1 &	9 - 13 &	$69.23\%$\\
Vibrant 2 &	19 - 38 &	$50.00\%$\\
  \bottomrule
\end{tabular}
\end{table}

\subsection{Hybrid Video Tests}
We conducted one experiment using a long hybrid video (Table~\ref{tab:hybrid}). The total amount of scenes in this experiment is 106. The overall accuracy of 2SDS is 99.06\%. Theoretically, the result of this experiment should sit between the previous two tests, however, an anomaly has arisen most likely due to the lack of samples. A more detailed experiment should be conducted to further determine the accuracy of 2SDS on hybrid videos.

\begin{table}
  \caption{Hybrid video test result.}
  \label{tab:hybrid}
  \begin{tabular}{lcr}
    \toprule
    Experiment No.&	Output - Truth&	Accuracy\\
    \midrule
    Hybrid 1 & 105 - 106 &	$99.06\%$\\
  \bottomrule
\end{tabular}
\end{table}

\section{Bringing in Spatial Information}
Bringing in spatial information and modeling techniques can potentially play a huge role in video interpretation. Previously difficult and untouchable problems like continuous gesture recognition and scene recognition are being cracked using the CNN-based spatio-temporal reasoning model \cite{DBLP:journals/corr/abs-1909-05165} and the 2SDS algorithm as well.

Our work has only utilised the temporal information in video stream, our future work can make use of graphs, and spatially model a frame into a graph, with the objects as the vertices and the spatial relations between the objects as the edges, like the MST-GNN \cite{DBLP:journals/tip/LiCZZWT21} and the VRD-GCN \cite{DBLP:conf/mm/QianZLXP019}. Doing this, we can extract even more information out of a video. For example, a person's gesture in a scene can be identified, and the scenes with more significant camera or object movements (e.g., the vibrant and hybrid video tests) will not cause significant problem for the algorithm because the spatial relation of the objects stays the same.

This future work would bring immense potential with the use of spatial information, which will add a whole other dimension of usable information that can benefit video analysis with richer semantics and the ability of grouping fast-moving frames, bringing video interpretation models yet another step closer to how human perceive visual information.

\section{Conclusion}
Under the context of real-time video stream analysis using temporal segmentation methods, we devised 2SDS, a temporal segmentation algorithm that can be used alongside CNNs to complement for the lack of temporal information handling ability of the CNN-based models. We gave, yet another powerful tool that CNN models can utilise, the ability to take advantage of the inherent temporal aspect of videos. Video stream analysis is a completely different task compared to image recognition, and we are finally seeing some evidence that we can still use 2D CNNs to interpret video information.

The 2SDS algorithm utilise a refined difference hash value method and a novel data smoothing and data selection technique to crack the temporal segmentation problem. Although there are still drawbacks with fast-moving frames in vibrant videos, the 2SDS algorithm has already done a great job at separating relatively simple and stationary scenes in videos, and it gets the job done at a respectful speed, which will allow the 2SDS to get a finer temporal resolution compared with neural networks.

For future work, some improvements on 2SDS (e.g., adding graphs to model spatial relations) can potentially boost the algorithm’s performance on fast-moving scenes and smooth transitions.

\begin{acknowledgments}
  We would like to dedicate our thank to Dr Zhiguo Long, Dr John Stell, and Dr Liu Yang, who have been extremely generous and helpful throughout the course of our work.  
\end{acknowledgments}

\bibliography{main-paper}

\begin{thebibliography}{11}
\expandafter\ifx\csname natexlab\endcsname\relax\def\natexlab#1{#1}\fi
\providecommand{\url}[1]{\texttt{#1}}
\providecommand{\href}[2]{#2}
\providecommand{\path}[1]{#1}
\providecommand{\DOIprefix}{doi:}
\providecommand{\ArXivprefix}{arXiv:}
\providecommand{\URLprefix}{URL: }
\providecommand{\Pubmedprefix}{pmid:}
\providecommand{\doi}[1]{\href{http://dx.doi.org/#1}{\path{#1}}}
\providecommand{\Pubmed}[1]{\href{pmid:#1}{\path{#1}}}
\providecommand{\bibinfo}[2]{#2}
\ifx\xfnm\relax \def\xfnm[#1]{\unskip,\space#1}\fi
\bibitem[{Stappen et~al.(2021)Stappen, Baird, Cambria, and
  Schuller}]{DBLP:journals/expert/StappenBCSC21}
\bibinfo{author}{L.~Stappen}, \bibinfo{author}{A.~Baird},
  \bibinfo{author}{E.~Cambria}, \bibinfo{author}{B.~W. Schuller},
\newblock \bibinfo{title}{Sentiment analysis and topic recognition in video
  transcriptions},
\newblock \bibinfo{journal}{{IEEE} Intell. Syst.} \bibinfo{volume}{36}
  (\bibinfo{year}{2021}) \bibinfo{pages}{88--95}.
\bibitem[{Patel et~al.(2022)Patel, Vyas, Vyas, and
  Ojha}]{DBLP:journals/mta/PatelV0O22}
\bibinfo{author}{A.~S. Patel}, \bibinfo{author}{R.~Vyas},
  \bibinfo{author}{O.~P. Vyas}, \bibinfo{author}{M.~Ojha},
\newblock \bibinfo{title}{A study on video semantics; overview, challenges, and
  applications},
\newblock \bibinfo{journal}{Multim. Tools Appl.} \bibinfo{volume}{81}
  (\bibinfo{year}{2022}) \bibinfo{pages}{6849--6897}.
\bibitem[{Pal et~al.(2021)Pal, Sekh, Dogra, Kar, Roy, and
  Prasad}]{DBLP:journals/csur/PalSDKRP21}
\bibinfo{author}{R.~Pal}, \bibinfo{author}{A.~A. Sekh}, \bibinfo{author}{D.~P.
  Dogra}, \bibinfo{author}{S.~Kar}, \bibinfo{author}{P.~P. Roy},
  \bibinfo{author}{D.~K. Prasad},
\newblock \bibinfo{title}{Topic-based video analysis: {A} survey},
\newblock \bibinfo{journal}{{ACM} Comput. Surv.} \bibinfo{volume}{54}
  (\bibinfo{year}{2021}) \bibinfo{pages}{118:1--118:34}.
\bibitem[{Diba et~al.(2017)Diba, Fayyaz, Sharma, Karami, Arzani, Yousefzadeh,
  and Gool}]{DBLP:journals/corr/abs-1711-08200}
\bibinfo{author}{A.~Diba}, \bibinfo{author}{M.~Fayyaz},
  \bibinfo{author}{V.~Sharma}, \bibinfo{author}{A.~H. Karami},
  \bibinfo{author}{M.~M. Arzani}, \bibinfo{author}{R.~Yousefzadeh},
  \bibinfo{author}{L.~V. Gool},
\newblock \bibinfo{title}{Temporal 3d convnets: New architecture and transfer
  learning for video classification},
\newblock \bibinfo{journal}{CoRR} \bibinfo{volume}{abs/1711.08200}
  (\bibinfo{year}{2017}). \href{http://arxiv.org/abs/1711.08200}{{\tt
  arXiv:1711.08200}}.
\bibitem[{Feichtenhofer et~al.(2019)Feichtenhofer, Fan, Malik, and
  He}]{DBLP:conf/iccv/Feichtenhofer0M19}
\bibinfo{author}{C.~Feichtenhofer}, \bibinfo{author}{H.~Fan},
  \bibinfo{author}{J.~Malik}, \bibinfo{author}{K.~He},
\newblock \bibinfo{title}{Slowfast networks for video recognition},
\newblock in: \bibinfo{booktitle}{2019 {IEEE/CVF} International Conference on
  Computer Vision, {ICCV} 2019, Seoul, Korea (South), October 27 - November 2,
  2019}, \bibinfo{publisher}{{IEEE}}, \bibinfo{year}{2019}, pp.
  \bibinfo{pages}{6201--6210}.
\bibitem[{Hochreiter and Schmidhuber(1997)}]{DBLP:journals/neco/HochreiterS97}
\bibinfo{author}{S.~Hochreiter}, \bibinfo{author}{J.~Schmidhuber},
\newblock \bibinfo{title}{Long short-term memory},
\newblock \bibinfo{journal}{Neural Comput.} \bibinfo{volume}{9}
  (\bibinfo{year}{1997}) \bibinfo{pages}{1735--1780}.
\bibitem[{Krawetz(2013)}]{Krawetz}
\bibinfo{author}{N.~Krawetz}, \bibinfo{title}{Kind of like that},
  \bibinfo{year}{2013}. \URLprefix
  \url{http://www.hackerfactor.com/blog/?/archives/529-Kind-of-Like-That.html}.
\bibitem[{Xu et~al.(2018)Xu, Yang, Fan, Yue, Liang, Yang, and
  Huang}]{DBLP:journals/corr/abs-1809-03327}
\bibinfo{author}{N.~Xu}, \bibinfo{author}{L.~Yang}, \bibinfo{author}{Y.~Fan},
  \bibinfo{author}{D.~Yue}, \bibinfo{author}{Y.~Liang},
  \bibinfo{author}{J.~Yang}, \bibinfo{author}{T.~S. Huang},
\newblock \bibinfo{title}{{YouTube-VOS}: {A} large-scale video object
  segmentation benchmark},
\newblock \bibinfo{journal}{CoRR} \bibinfo{volume}{abs/1809.03327}
  (\bibinfo{year}{2018}). \href{http://arxiv.org/abs/1809.03327}{{\tt
  arXiv:1809.03327}}.
\bibitem[{K{\"{o}}p{\"{u}}kl{\"{u}} et~al.(2019)K{\"{o}}p{\"{u}}kl{\"{u}},
  Herzog, and Rigoll}]{DBLP:journals/corr/abs-1909-05165}
\bibinfo{author}{O.~K{\"{o}}p{\"{u}}kl{\"{u}}}, \bibinfo{author}{F.~Herzog},
  \bibinfo{author}{G.~Rigoll},
\newblock \bibinfo{title}{Comparative analysis of {CNN}-based spatiotemporal
  reasoning in videos},
\newblock \bibinfo{journal}{CoRR} \bibinfo{volume}{abs/1909.05165}
  (\bibinfo{year}{2019}). \href{http://arxiv.org/abs/1909.05165}{{\tt
  arXiv:1909.05165}}.
\bibitem[{Li et~al.(2021)Li, Chen, Zhao, Zhang, Wang, and
  Tian}]{DBLP:journals/tip/LiCZZWT21}
\bibinfo{author}{M.~Li}, \bibinfo{author}{S.~Chen}, \bibinfo{author}{Y.~Zhao},
  \bibinfo{author}{Y.~Zhang}, \bibinfo{author}{Y.~Wang},
  \bibinfo{author}{Q.~Tian},
\newblock \bibinfo{title}{Multiscale spatio-temporal graph neural networks for
  {3D} skeleton-based motion prediction},
\newblock \bibinfo{journal}{{IEEE} Trans. Image Process.} \bibinfo{volume}{30}
  (\bibinfo{year}{2021}) \bibinfo{pages}{7760--7775}.
\bibitem[{Qian et~al.(2019)Qian, Zhuang, Li, Xiao, Pu, and
  Xiao}]{DBLP:conf/mm/QianZLXP019}
\bibinfo{author}{X.~Qian}, \bibinfo{author}{Y.~Zhuang},
  \bibinfo{author}{Y.~Li}, \bibinfo{author}{S.~Xiao}, \bibinfo{author}{S.~Pu},
  \bibinfo{author}{J.~Xiao},
\newblock \bibinfo{title}{Video relation detection with spatio-temporal graph},
\newblock in: \bibinfo{editor}{L.~Amsaleg}, \bibinfo{editor}{B.~Huet},
  \bibinfo{editor}{M.~A. Larson}, \bibinfo{editor}{G.~Gravier},
  \bibinfo{editor}{H.~Hung}, \bibinfo{editor}{C.~Ngo}, \bibinfo{editor}{W.~T.
  Ooi} (Eds.), \bibinfo{booktitle}{Proceedings of the 27th {ACM} International
  Conference on Multimedia, {MM} 2019, Nice, France, October 21-25, 2019},
  \bibinfo{publisher}{{ACM}}, \bibinfo{year}{2019}, pp.
  \bibinfo{pages}{84--93}.

\end{thebibliography}

\end{document}